\documentclass{article}

\PassOptionsToPackage{numbers, compress}{natbib}

\usepackage[final]{neurips_bdl2019}

\usepackage[utf8]{inputenc} 
\usepackage[T1]{fontenc}    
\usepackage{hyperref}       
\usepackage{url}            
\usepackage{booktabs}       
\usepackage{amsfonts}       
\usepackage{nicefrac}       
\usepackage{microtype}      

\hypersetup{}

\usepackage{microtype}
\usepackage{caption}
\usepackage{subcaption}
\usepackage{booktabs} 
\usepackage{tabularx}

\usepackage[T1]{fontenc}    
\usepackage{amsfonts}       
\usepackage{amssymb}
\usepackage{amsmath}
\usepackage{nicefrac}       

\usepackage[utf8]{inputenc}

\usepackage[pdftex]{graphicx}
\graphicspath{{images/}}
\DeclareGraphicsExtensions{.pdf,.jpeg,.png}

\usepackage{algorithm}
\usepackage{algorithmic}
\usepackage{url}
\usepackage{wrapfig}

\usepackage{natbib}

\usepackage{tikz}
\usetikzlibrary{bayesnet}

\usepackage{environ}
\makeatletter
\newsavebox{\measure@tikzpicture}
\NewEnviron{scaletikzpicturetowidth}[1]{%
  \def\tikz@width{#1}%
  \begin{lrbox}{\measure@tikzpicture}%
  \BODY
  \end{lrbox}%
  \pgfmathparse{#1/\wd\measure@tikzpicture}%
  \BODY
}
\makeatother

\usepackage{xr}

\makeatletter
\newcommand*{\addFileDependency}[1]{
  \typeout{(#1)}
  \@addtofilelist{#1}
  \IfFileExists{#1}{}{\typeout{No file #1.}}
}
\makeatother

\newcounter{docpart}

\newcounter{olddocpart}

\usepackage{wrapfig}

\usepackage{multirow}

\newcommand{\bs}{\boldsymbol}

\newcommand{\x}{{\bs{x}}}
\renewcommand{\xi}{\x^{i}}

\newcommand{\z}{{\bs z}}

\newcommand{\params}{{\bs \theta}}

\newcommand{\M}{\mathcal{M}}
\newcommand{\Mmodel}{\M_{\params}}
\newcommand{\Msamp}{\M_{\mathcal{S}}}

\newcommand{\pjoint}{\mathcal{P}}

\newcommand{\pdec}{p}
\newcommand{\penc}{q}

\newcommand{\Mdec}{\pdec_{\params}}
\newcommand{\Menc}{\penc_{\params}}

\newcommand{\Mencoder}{\Menc(\x | \z)}
\newcommand{\Mdecoder}{\Mdec(\x | \z)}

\newcommand{\Pencoderjoint}{\Menc(\x , \z)}
\newcommand{\Pdecoderjoint}{\Mdec(\x , \z)}

\newcommand{\Penc}{\Menc(\z | \x)\, \pjoint(\x)}
\newcommand{\Pdec}{\Mdec(\x | \z)\, \pjoint(\z)}
\newcommand{\Mdecjoint}{\Mdec \left(\x, \z \right)}
\newcommand{\Mencjoint}{\Menc \left(\x, \z \right)}

\newcommand{\dz}{\mathrm{d}\z}

\newcommand{\MIM}{{MIM }}
\newcommand{\VAEloss}{\mathcal{L}_\text{VAE}}

\newcommand{\MIMloss}{\mathcal{L}_\text{MIM}}

\newcommand{\EMIMloss}{\hat{\mathcal{L}}_\text{MIM}}
\newcommand{\CEloss}{\mathcal{L}_\text{CE}}

\newcommand{\DKL}[2]{\mathcal{D}_\text{KL}\left(#1\,\|\, #2\right)}

\newcommand{\E}[2]{\mathbb{E}_{#1}\left[#2\right]}

\newcommand{\RMIM}{\mathrm{R}_{\mathrm{MIM}}}
\newcommand{\RH}{\mathrm{R}_\mathrm{H}}

\newcommand{\eg}{{\em e.g.}}
\newcommand{\ie}{{\em i.e.}}

\title{High Mutual Information in Representation Learning with Symmetric Variational Inference}

\author{%
  Micha Livne \\
  University of Toronto\\
  Vector Institute \\
  \texttt{mlivne@cs.toronto.edu} \\
   \And
   Kevin Swersky \\
   Google Research \\
   \texttt{kswersky@google.com} \\
   \And
   David J.\ Fleet \\
  University of Toronto\\
  Vector Institute \\
   \texttt{fleet@cs.toronto.edu} \\
}

\begin{document}

\vspace*{-1cm}
\maketitle
\vspace*{-0.75cm}

\begin{abstract}
    We introduce the Mutual Information Machine (MIM), a novel formulation of representation learning, using a joint distribution over the observations and latent state in an encoder/decoder framework. Our key principles are symmetry and mutual information, where symmetry encourages the encoder and decoder to learn different factorizations of the same underlying distribution, and mutual information, to encourage the learning of useful representations for downstream tasks. Our starting point is the symmetric Jensen-Shannon divergence between the encoding and decoding joint distributions, plus a mutual information encouraging regularizer. We show that this can be bounded by a tractable cross entropy loss function between the true model and a parameterized approximation, and relate this to the maximum likelihood framework. We also relate MIM to variational autoencoders (VAEs) and demonstrate that MIM is capable of learning symmetric factorizations, with high mutual information that avoids posterior collapse.
\end{abstract}

\vspace*{-0.2cm}
\section{Introduction}
\label{sec:introduction}
\vspace*{-0.2cm}

The variational auto-encoder (VAE) \cite{Kingma2013} is a class of hierarchical Bayesian models based on transforming data from a simple latent prior into a complicated observation distribution using a neural network. The model density is represented by the marginal distribution.
$\int_\x \Pdec \dz$, 
where $\params$ denotes the model parameters. 
The data log-likelihood is in general intractable to compute, 
so VAEs maximize the so-called evidence lower bound (ELBO).
\begin{equation}
    \log \pjoint (\x) ~\ge~ 
    \E{\z \sim \Menc(\z|\x)}{\,\log \Mdec(\x|\z)\,} - \DKL{\Menc(\z|\x)}{\pjoint(\z)} ~ , \label{eq:elbo}
\end{equation}
where $\Penc$ denotes an approximate posterior that maps observations to a distribution over latent variables, and where $\Pdec$ denotes the mapping from latent codes to observations. While VAEs have been successfully applied in a number of settings \cite{Germain2015,Klys2018,Kingma2016,Berg2018}, they suffer from a shortcoming known as posterior collapse \cite{DBLP:journals/corr/BowmanVVDJB15,ChenKSDDSSA16,DBLP:journals/corr/abs-1901-03416,DBLP:journals/corr/OordKK16,DBLP:journals/corr/abs-1711-00937}. Effectively, the penalty on the prior term in Eq.\ \eqref{eq:elbo} causes the encoder to lose information in some dimensions of the latent code, leading the decoder to ignore those dimensions. The result is a model that generates good samples, but suffers from a poor latent representation. This is an issue for any downstream application that relies on $\z$.

Here we present a related but distinct framework called Mutual Information Machine (MIM). Our objectives are twofold: to promote symmetry in the encoding and decoding distributions (\ie, consistency), and to encourage high mutual information between $\x$ and $\z$ (\ie, good representation). By symmetry, we mean that the encoding and decoding distributions should represent two equivalent factorizations of the same underlying joint distribution. The framework can be seen as a symmetric analogue of VAEs, where we optimize a symmetric divergence instead of the asymmetric KL. In preliminary experiments, we find that MIM produces highly informative representations, with comparable sample quality to VAEs.

\vspace*{-0.2cm}
\section{Mutual Information Machine}
\label{sec:mim}
\vspace*{-0.2cm}

To begin, it is helpful to write down the VAE loss function as the KL divergence between two joint distributions over $\x$ and $\z$, as in~\cite{pu2017adversarial} i.e., up to an additive constant the following holds.
\begin{equation}
    \VAEloss \left( \params \right) ~=~ \DKL{ \Menc(\z | \x)\pjoint(\x)}{ \Mdec(\x | \z) \pjoint(\z) } ~.
    \label{eq:vae-kl}
\end{equation}
This loss could be symmetrized by using the symmetric KL divergence, however this would result in a $\log \pjoint(\x)$ term that cannot be evaluated, as this distribution does not typically have an analytical form. In~\citet{pu2017adversarial}, minimization of the symmetric KL loss is expressed as a stationary point of an adversarial min-max objective.

Here we propose to optimize a bound on a regularized version of the Jensen-Shannon divergence (JSD). 
Defining the mixture distribution $\Msamp \equiv \frac{1}{2} \big( \,\Pdec + \Penc \, \big)$, the 
JSD is defined as
\begin{equation}
\vspace*{-0.1cm}
    \mathrm{JSD}(\params) ~=~ \frac{1}{2}\Big( \, \DKL{\Pdec}{\Msamp} + \DKL{\Penc}{\Msamp}\, \Big) ~.
    \label{eq:jsd-h}
\end{equation}
A second principle of our formulation involves encouraging a representation with high mutual information. Mutual information is related to joint entropy by the identity $H(\x, \z)= H(\x) + H(\z) - I(\x; \z)$. Since $H(\x, \z)$ is tractable up to a constant under both the encoding and decoding distributions, we add the regularizer\footnote{We use $H_q(\x,\z)$ and $MI_q(\x;\z)$ to notate entropy and mutual information under a distribution $q(\x,\z)$} $\RH(\params) = \frac{1}{2}(H_{\Menc(\z | \x)\pjoint(\x)}(\x, \z) + H_{\Mdec(\x | \z) \pjoint(\z)}(\x, \z))$ to $\mathrm{JSD}(\params)$, which can be shown to be equal to $H_{\Msamp}(\x, \z)$, the entropy of the mixture distribution $\Msamp$.

This formulation allows us to derive a tractable (in terms of being amenable to optimization via reparameterization) bound on the objective. First we define parameterized joint distributions,
$\Menc (\x, \z )  \equiv \Menc\left(\z| \x\right) \Menc\left(\x\right)$, and $\Mdec (\x, \z )  \equiv \Mdec\left(\x| \z\right) \Mdec\left(\z\right)$.
Where $\Mdec\left(\z\right)$ can be parameterized for added flexibility, or set to $\pjoint(\z)$. $\Menc(\x)$ is an observation model which could be e.g., a normalizing flow~\cite{dinh2016density, kingma2018glow} or an auto-regressive model~\cite{van2016conditional}. For a fair comparison with VAEs, we leverage the encoder and decoder distributions $\Mencoder$, $\Mdecoder$ to avoid adding additional parameters. Details are given in the supplementary material.

Defining $\Mmodel \equiv \frac{1}{2}(\Menc (\x, \z ) + \Mdec (\x, \z ))$, the bound can be expressed as,
\begin{equation}
\vspace*{-0.1cm}
    H_{\Msamp}(\x, \z)
    \,\leq\,  H(\Msamp, \Mmodel) 
    \,\leq\, \frac{1}{2}\big(\, H(\Msamp, \Menc ) + H(\Msamp, \Mdec ) \, \big) \,\equiv\, \MIMloss(\params)
    \label{eq:mim-loss}
\vspace*{-0.1cm}
\end{equation}
where $H(p, q)$ is the cross entropy between distributions $p$ and $q$. While the first inequality no longer depends on $\log \pjoint(\x)$ and is therefore tractable, the second inequality, which follows from Jensen's inequality, can be shown to additionally encourage consistency in $\Mmodel$ (\ie, $\Menc (\x, \z )=\Mdec (\x, \z )$). Further, when $\Menc\left(\x\right)=\pjoint(\x)$ and $\Mdec\left(\z\right)=\pjoint(\z)$, it can be shown that $\MIMloss(\params)$ is equal to the symmetric KL loss plus the regularizer $\RH(\params)$. See supplementary material for more details.

We have therefore derived a loss function that encourages both symmetry in the encoding and decoding distributions, as well as high mutual information in the learned representation. Further, this loss function can be directly minimized without requiring an adversarial reformulation. In the experiments, we will show preliminary results exploring the properties of MIM.


\vspace*{-0.2cm}
\section{Experiment: 2D Mixture Model Data} 
\label{sec:posterior-collapse-mim-vae}
\vspace*{-0.2cm}

\begin{figure}[t]
    \centering
    \setlength{\tabcolsep}{0pt}
    \begin{tabular}{*6{>{\centering\arraybackslash}m{0.167\textwidth}}}
      \includegraphics[width=0.165\columnwidth]{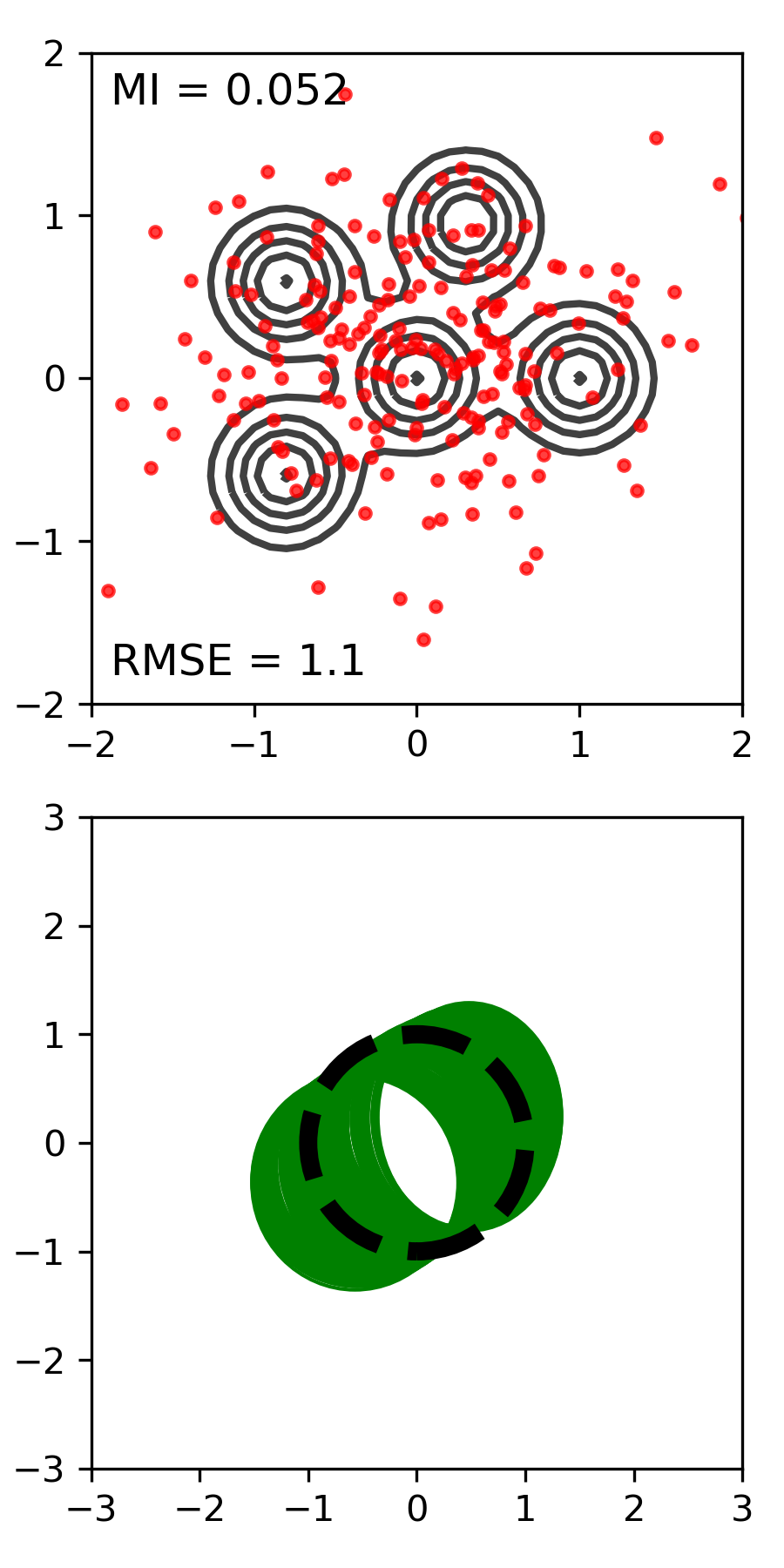}
    & \includegraphics[width=0.165\columnwidth]{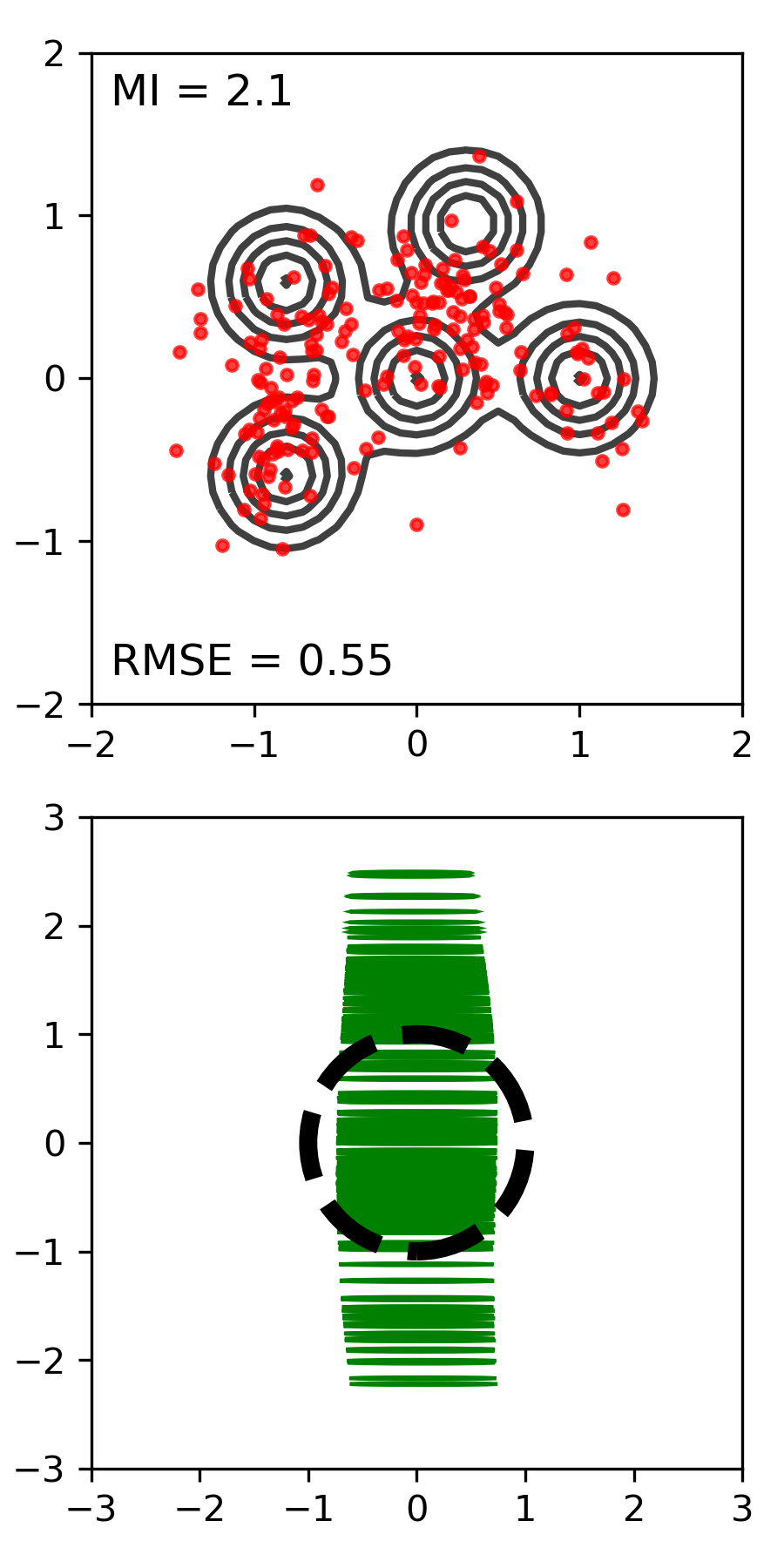}
    & \includegraphics[width=0.165\columnwidth]{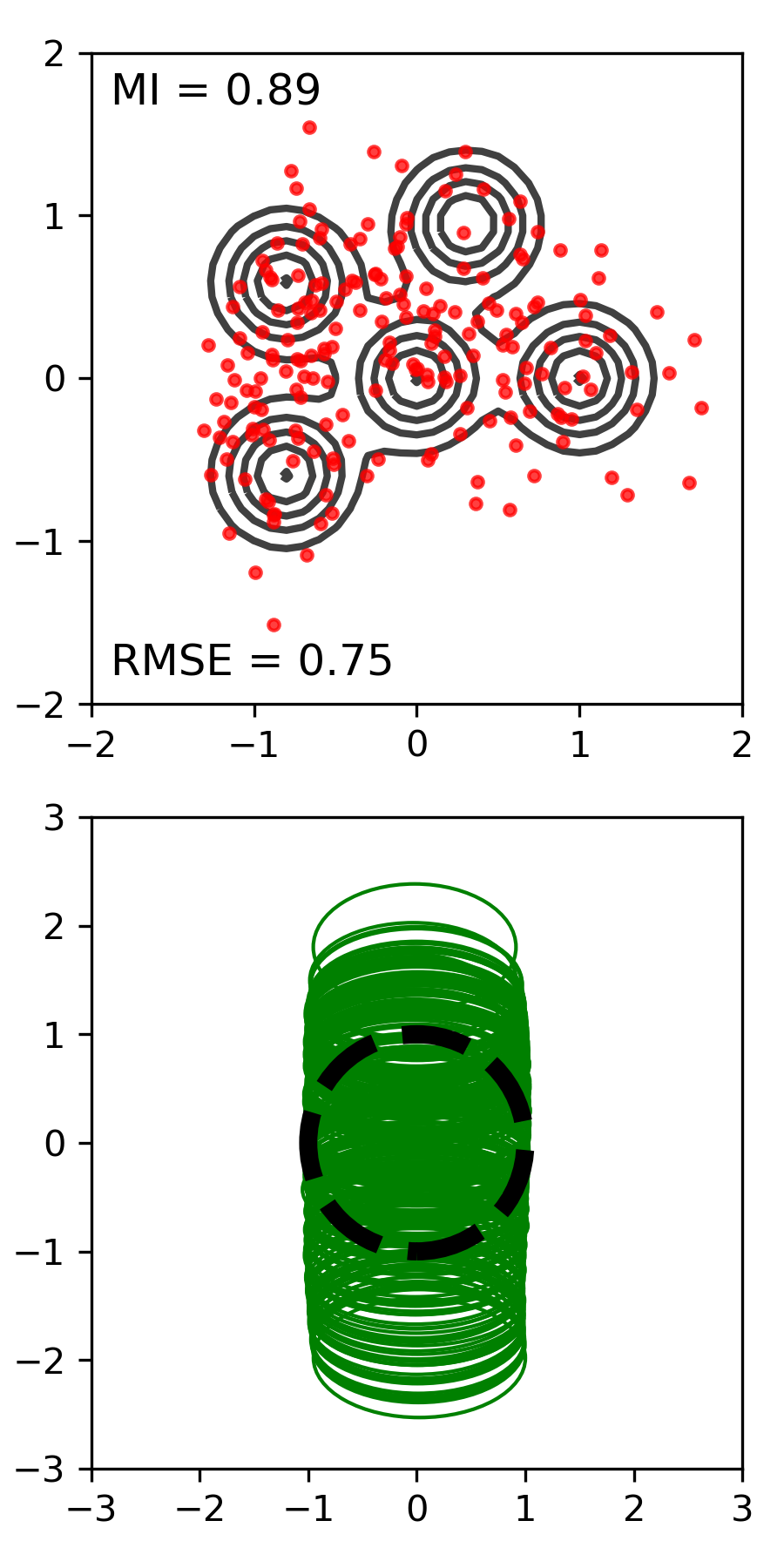}
    & \includegraphics[width=0.165\columnwidth]{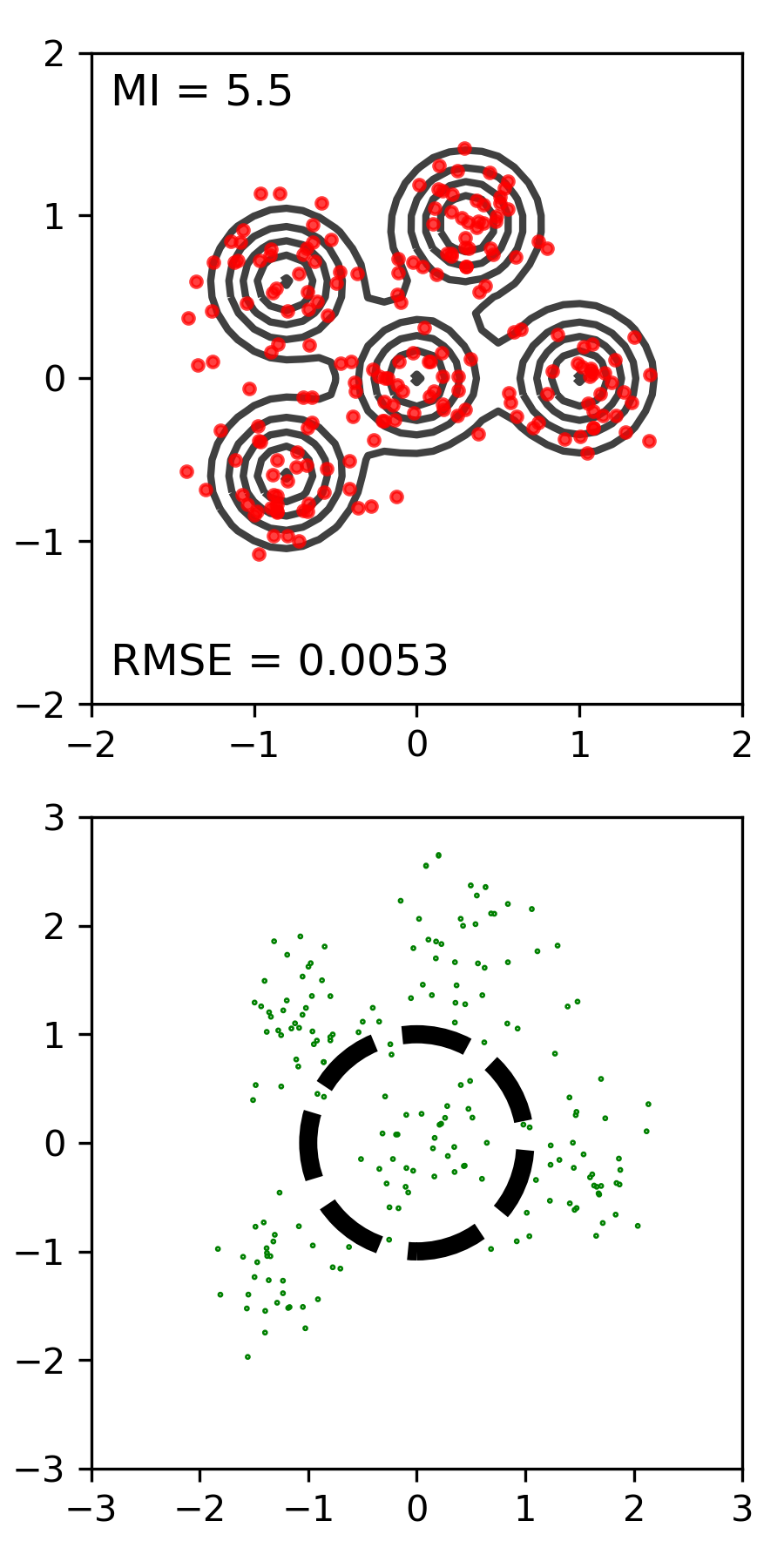}
    & \includegraphics[width=0.165\columnwidth]{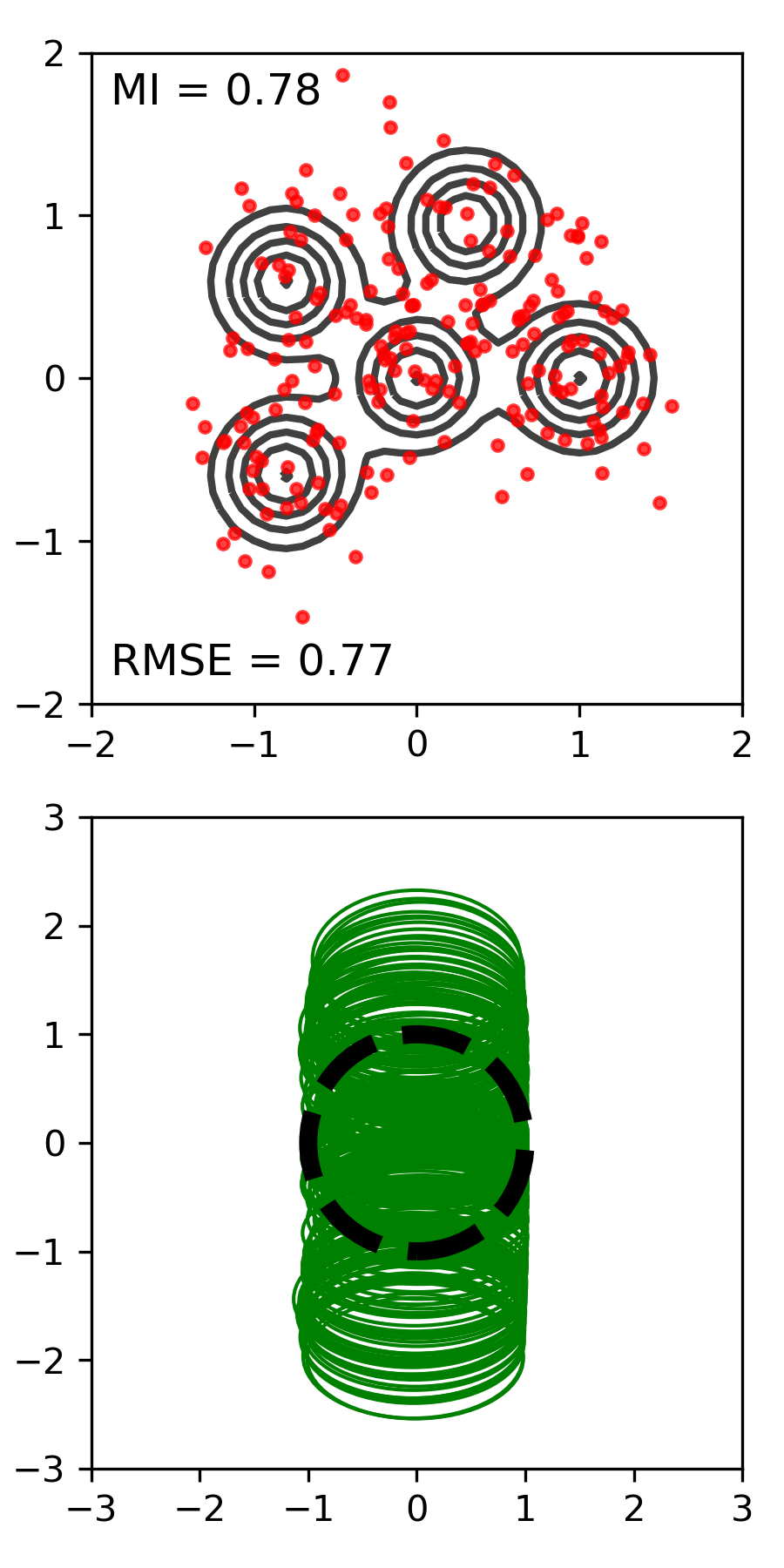}
    & \includegraphics[width=0.165\columnwidth]{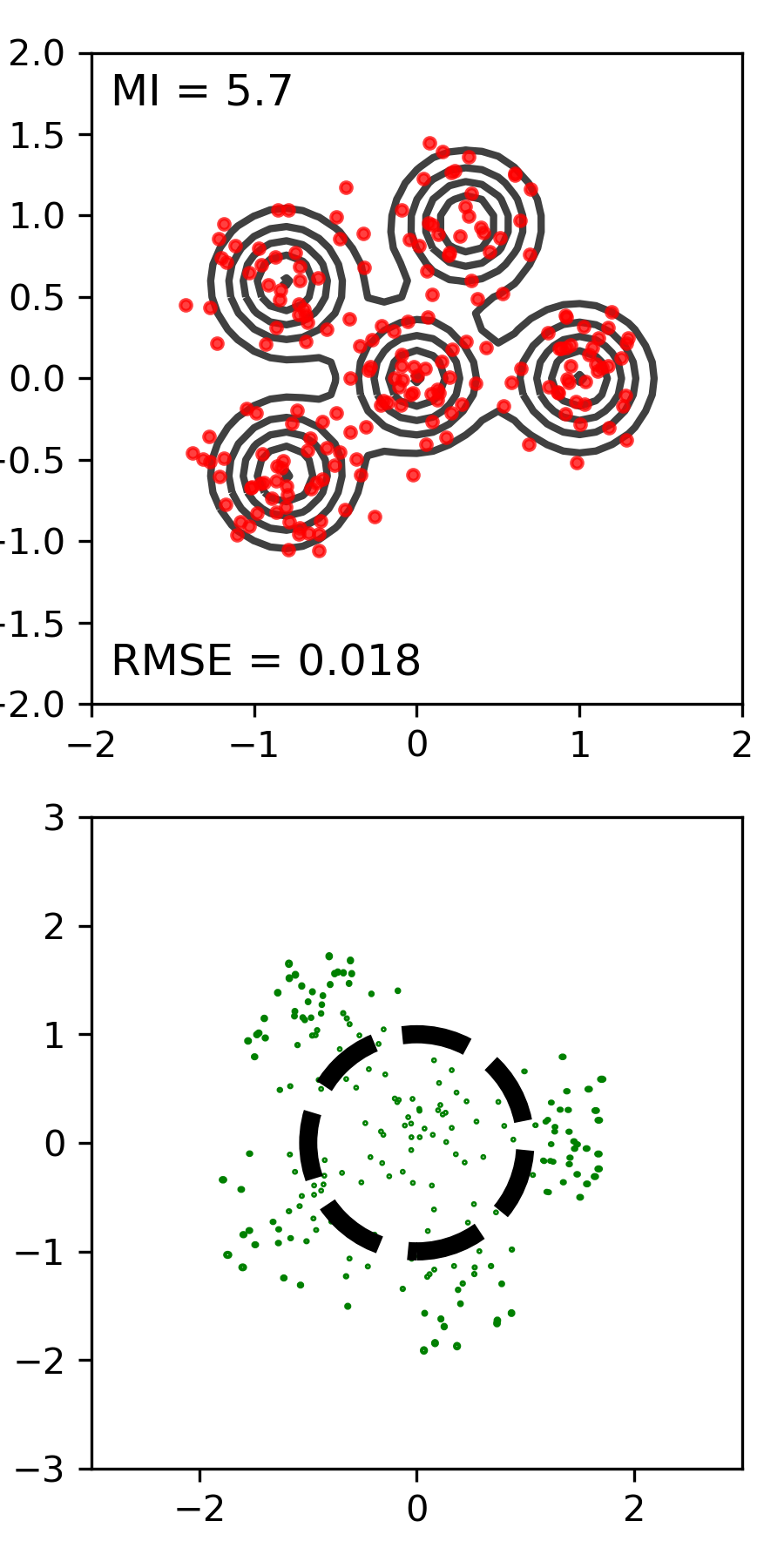}
    \\
    \multicolumn{2}{c}{(a) $h \in \mathbb{R}^{5}$ } & \multicolumn{2}{c}{(b) $h \in \mathbb{R}^{20}$ } & \multicolumn{2}{c}{(c) $h \in \mathbb{R}^{500}$ } \\
    \end{tabular}
    \caption{
    VAE (odd columns) and MIM (even columns) with 2D inputs, a 2D latent space, and 5, 20 and 500 hidden units. 
    Top: Black contours are level sets of data distribution $\pjoint(\x)$, red points are 
    reconstructed samples drawn from the decoder $\Mdec(\x|\z')$, where $\z' \sim \Menc(\z | \x')$ for
    data point $\x'$ from $\pjoint(\x)$.
    Bottom: The dashed black circle depicts one standard deviation of $\pjoint(\z)$. Each green 
    curve depicts a one standard deviation ellipse of the encoder posterior $\Menc(\z | \x')$.
    (a) For weak architectures MIM and VAE exhibit high posterior variance.
    (b,c) For more expressive architectures the VAE predictive variance remains high,
    an indication of posterior collapse.
    MIM generally produces lower predictive variance and lower reconstruction 
    errors, consistent with high mutual information.
    }\label{fig:posterior-collapse-qualitative}
    \vspace*{-0.5cm}
\end{figure}


We begin with a dataset of 2D observations $\x \in \mathbb{R}^2$
drawn from a Gaussian mixture model, and a 2D latent space, $\z \in \mathbb{R}^2$. 
In 2D we can easily visualize the model and measure quantitative properties 
of interest (\eg, mutual information). 
(Complete experiment details given in supplementary material).

Figure \ref{fig:posterior-collapse-qualitative} depicts results 
for the VAE (even columns) and MIM (odd columns) using single-layer encoder and decoder networks, with increasing numbers 
of hidden units (moving left to right) to control model expressiveness.
The top row (for VAE and MIM respectively) depicts observation 
space. With each case we also report the mutual information and the root-mean-squared observation reconstruction error when sampling the predictive encoder/decoder distributions, with MIM showing a superior performance. 
See additional results in the supplementary material (Fig.\ \ref{fig:posterior-collapse-quantitative}).

The bottom row of Fig.\ \ref{fig:posterior-collapse-qualitative}
depicts the latent space behavior. For the weakest architecture, with only 5 
hidden units, both MIM and VAE posteriors have large variances.
When the number of hidden units increases, however, it is 
clear that while the VAE posterior variance remains very large in
one dimension (\ie, a common sign of posterior collapse), the MIM encoder produces much tighter posteriors 
densities, which capture the global (\ie, aggregated) structure of the observations.  

In addition, with a more expressive architecture, i.e., more hidden units, the MIM 
encoding variance is extremely small, and the reconstruction error approaches 0.
In effect, the encoder and decoder learn an (approximately) invertible mapping using 
an unconstrained architecture (demonstrated here in 2D), when the dimensionality of the 
latent representation and the observations is the same.

The VAE, by comparison, is prone to posterior collapse, reflected in relatively low mutual information. 
In this regard, we note that several 
papers have described ways to mitigate posterior collapse in VAE learning, e.g., 
by lower bounding or annealing the KL divergence term in the VAE objective 
(e.g., \citep{DBLP:journals/corr/abs-1711-00464,DBLP:journals/corr/abs-1901-03416}), 
or by limiting the expressiveness of the decoder (e.g., \citep{ChenKSDDSSA16}).
We posit that MIM does not suffer from this problem as a consequence of 
the objective design principles that encourage high mutual information
between observations and the latent representation.

\vspace*{-0.2cm}
\section{Experiment: MIM Representations with High Dimensional Image Data} 
\label{sec:representation-learning-with-mim}
\vspace*{-0.5cm}

\begin{minipage}{\textwidth}
    \begin{minipage}[b]{0.49\textwidth}
    \centering
    \setlength{\tabcolsep}{0pt}
    \begin{tabular}{*4{>{\centering\arraybackslash}m{0.25\textwidth}}}
    \includegraphics[width=0.25\columnwidth]{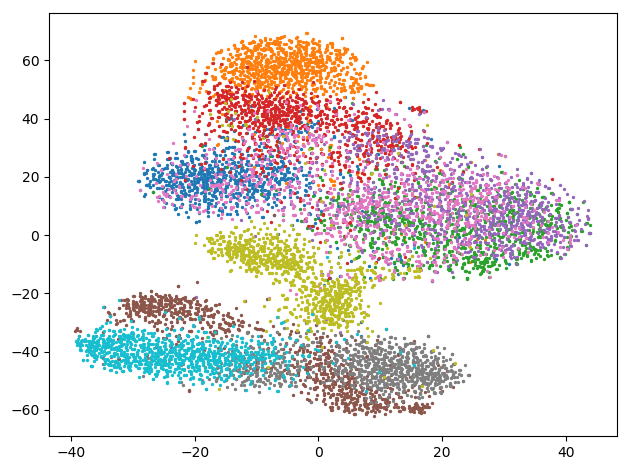}
    & \includegraphics[width=0.25\columnwidth]{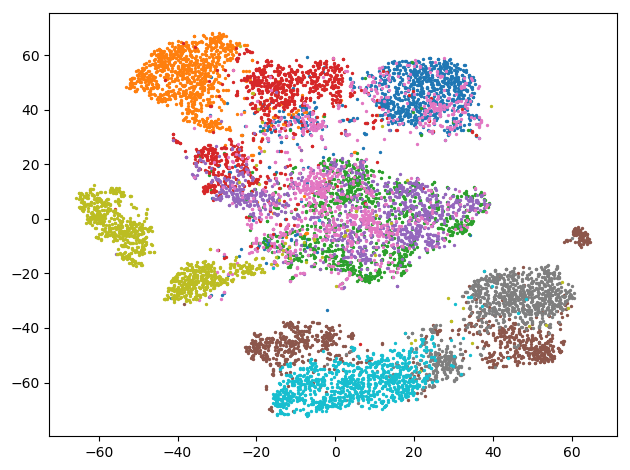} 
    & \includegraphics[width=0.25\columnwidth]{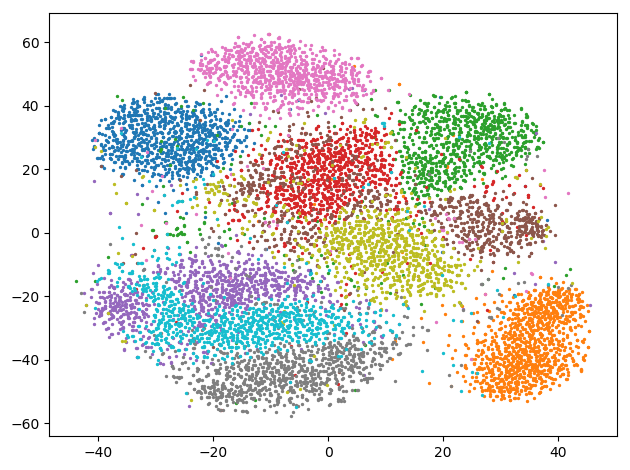}
    & \includegraphics[width=0.25\columnwidth]{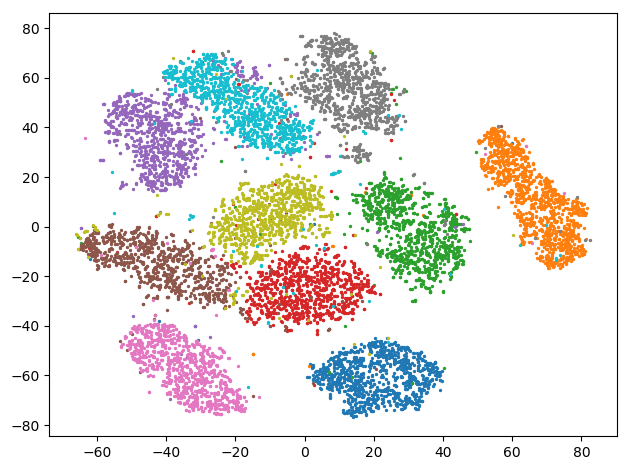} 
    \\
    \multicolumn{2}{c}{Fashion-MNIST} &  \multicolumn{2}{c}{MNIST}   \\
    \end{tabular}
    \captionof{figure}{VAE (odd) and MIM (even) $\z$ embedding for Fashion MNIST and MNIST with convHVAE (S) architecture (see Table 1).}
    \label{fig:mim-vs-vae-image-z-embed}
    \end{minipage}
    \hfill
    \begin{minipage}[b]{0.49\textwidth}
    \centering
    \setlength{\tabcolsep}{0.5em} 
    {
    \scriptsize
    \renewcommand{\arraystretch}{1.2}
    \begin{tabular}{l||c|c||c|c||}
         \multicolumn{1}{l||}{} &  \multicolumn{2}{c||}{convHVAE (S)}  & \multicolumn{2}{c||}{convHVAE (VP)}  \\
         Dataset & {\tiny MIM}  & {\tiny VAE} & {\tiny MIM}  & {\tiny VAE}   \\
        \hline
        \multirow{1}{*}{Fashion MNIST}
        & \textbf{0.84}  & 0.75 & \textbf{0.84}  & 0.76   \\
        \hline
        \multirow{1}{*}{MNIST}
        & \textbf{0.98}  & 0.91  & \textbf{0.98}  & 0.85  \\
    \end{tabular}
    }
    \vspace*{0.25cm}
    \captionof{table}{Test classification accuracy of 5-NN classifier for MIM and VAE learning. MIM shows superior clustering of classes in the latent representation in an unsupervised manner.}
    \label{tab:mim-vs-vae-image-quantitative-cls}
    \end{minipage}
\end{minipage}

Here we explore learning on higher dimensional image data, where we cannot accurately estimate mutual information~\cite{Hjelm2018}. Instead, following \cite{hjelm2018learning}, we focus on an auxiliary classification task as a proxy for the quality of the learned representation and on qualitative visualization of it. We experiment with MNIST \cite{LeCun1998}, and Fashion MNIST \cite{DBLP:journals/corr/abs-1708-07747}.
In what follows we also explore multiple architectures of VAE models from \cite{DBLP:journals/corr/TomczakW17}, and the corresponding MIM models (see Algorithm.\ \ref{algo:vae-as-mim} in the supplementary material), where, again we use VAE as the baseline. See supplementary material for the  details.

For the auxiliary transfer learning classification task we opted for K-NN classification, being a non-parametric method which represents the clustering in the latent representation without any additional training. 
We show quantitative results in Table\ \ref{tab:mim-vs-vae-image-quantitative-cls} for K-NN classification ($k=5$). 
We also present the corresponding qualitative visual clustering results (\ie, projection to 2D using t-SNE \cite{maaten2008visualizing}) in Fig.\ \ref{fig:mim-vs-vae-image-z-embed}.
Here, it is clear that MIM learning tends to cluster classes in the latent representation better than VAE, for an identical parameterization of a model.

\bibliographystyle{plainnat}
\bibliography{paper}

\newpage
\appendix
\section{Detail Derivation of MIM Learning}

Here we provide a detailed derivation of the loss of MIM learning, as defined in \eqref{eq:mim-loss}.
We would like to formulate a loss function which includes 
\eqref{eq:jsd-h} that reflects our desire for model symmetry and high mutual information.
This objective is difficult to optimize directly since we do not know how to evaluate 
$\log\pjoint(\x)$ in the general case (\ie, we do not have an exact closed-form 
expression for $\pjoint(\x)$).
As a consequence, we introduce parameterized approximate priors, $\Menc(\x)$ and $\Mdec(\z)$,
to derive tractable bounds on the penalized Jensen-Shannon divergence.
This is similar in spirit to VAEs, which introduce a parameterized approximate posterior.
These parameterized priors, together with the conditional encoder and decoder,
$\Menc(\z | \x)$ and $\Mdec(\x | \z)$, comprise a new pair of joint distributions,
\begin{align*}
    \Menc (\x, \z ) & \equiv \Menc\left(\z| \x\right) \Menc\left(\x\right) \\
    \Mdec (\x, \z ) & \equiv \Mdec\left(\x| \z\right) \Mdec\left(\z\right)  ~.
\end{align*}

These new joint distributions allow us to formulate a new, tractable loss
that bounds $H(\Msamp)$:
\begin{eqnarray}
    \CEloss(\params)
    &\equiv& H(\Msamp, \Mmodel)   \nonumber \\
    &=&  \DKL{\Msamp}{\Mmodel} + H(\Msamp) \nonumber \\
    &\geq&  H(\Msamp) ~,
    \label{eq:LCE}
\end{eqnarray}
where $H(\Msamp, \Mmodel)$ denotes the cross-entropy between $\Msamp$ and $\Mmodel$,
and
\begin{equation}
    \Mmodel ~= ~ \frac{1}{2} \big( \, \Mdecjoint + \Mencjoint \, \big) ~.
\end{equation}
In what follows we refer to $\CEloss$ as the cross-entropy loss.
It aims to match the model prior distributions to the anchors, while also
minimizing $H(\Msamp)$. The main advantage of this formulation is that the
cross-entropy loss can be trained by Monte Carlo sampling from the anchor
distributions with the reparameterization trick~\citep{Kingma2013, Rezende2014}.

At this stage it might seem odd to introduce a parametric prior for $\pjoint(\z)$.
Indeed, setting it directly is certainly an option.
Nevertheless, in order to achieve consistency between $\Mdecjoint$ and $\Mencjoint$
it can be advantageous to allow $\Mdec(\z)$ to vary.
Essentially, we trade-off latent prior fidelity for increased model consistency.

One issue with $\CEloss$ is that, while it will try to enforce consistency
between the model and the anchored distributions, i.e.,
$\Mdec (\x, \z ) \approx \Mdec(\x | \z)\pjoint(\z)$ and
$\Menc(\x, \z) \approx \Menc(\z | \x)\pjoint(\x)$, it will not directly
try to achieve model consistency: $\Mdec (\x, \z ) \approx \Menc(\x, \z)$.
To remedy this, we bound $\CEloss$ using Jensen's inequality, \ie,
\begin{align}
    \MIMloss(\params)~ &\equiv ~\frac{1}{2}
    \big(\, H(\Msamp, \Menc \left(\x, \z \right) ) + H(\Msamp, \Mdec \left(\x, \z \right)) \, \big)
    \label{eq:mimloss}  \\
    &\geq ~\,  \CEloss(\params) ~.
    \label{eq:mim_celoss}
\end{align}

Equation \eqref{eq:mimloss} gives us the loss function for the Mutual Information 
Machine (MIM).  It is an average of cross entropy terms between the mixture 
distribution $\Msamp$ and the model encoding and decoding distributions respectively.
To see that this encourages model consistency, it can be shown that $\MIMloss$ is
equivalent to $\CEloss$ plus a non-negative model consistency regularizer; i.e.,
\begin{align}
    \MIMloss(\params) ~= ~\CEloss(\params) + \RMIM(\params) ~.
    \label{eq:LMIM-LCE}
\end{align}
The non-negativity of $\RMIM$ is a simple consequence of 
$\MIMloss(\params) \ge \CEloss(\params)$ in \eqref{eq:mim_celoss}.

In what follows we derive the form of the MIM consistency regularizer in Eq.\ \eqref{eq:mim-loss}, named $ \RMIM(\params)$.
Recall that we define $\Mmodel=\frac{1}{2}(\Pdecoderjoint + \Pencoderjoint)$.
We can show that $\MIMloss$ is equivalent to $H(\Msamp, \Mmodel)$ plus a regularizer by taking their difference.
\begin{align}
    \RMIM(\params) &= \MIMloss(\params) - H(\Msamp, \Mmodel) \label{eq:RMIM} \\
    &= \frac{1}{2}(H(\Msamp, \Pdecoderjoint) + H(\Msamp, \Pencoderjoint)) - H(\Msamp, \Mmodel) \nonumber \\
    &= \frac{1}{2}(\DKL{\Msamp}{\Pdecoderjoint} + H_{\Msamp}(\x, \z) + \DKL{\Msamp}{\Pencoderjoint} + H_{\Msamp}(\x, \z)) \nonumber \\
    & ~~~~~  - \DKL{\Msamp}{\Mmodel} - H_{\Msamp}(\x, \z) \nonumber \\
    &=  \frac{1}{2}(\DKL{\Msamp}{\Pdecoderjoint} + \DKL{\Msamp}{\Pencoderjoint}) - \DKL{\Msamp}{\Mmodel} \nonumber
\end{align}
where $\RMIM(\params)$ is non-negative, and is zero 
only when the two joint model distributions, $\Menc \left(\x, \z \right)$
and $\Mdec \left(\x, \z \right)$, are identical under fair samples from 
the joint sample distribution $\Msamp \left(\x, \z \right)$.
To prove that $\RMIM(\params) \ge 0$ we now construct Equation\ \eqref{eq:RMIM}
in terms of expectation over a joint distribution, which yields
\begin{align*}
    \RMIM(\params) &= \frac{1}{2}(H(\Msamp, \Pdecoderjoint) + H(\Msamp, \Pencoderjoint)) - H(\Msamp, \Mmodel) \\
    &= ~ \E{\x,\z \sim \Msamp}{-\frac{1}{2} \log \Pdecoderjoint - \frac{1}{2} \log \Pencoderjoint + \log \frac{1}{2} (\Menc \left(\x, \z \right) + \Mdec \left(\x, \z \right))} \\
    &= ~ \E{\x,\z \sim \Msamp}{- \log \sqrt{\Menc \left(\x, \z \right) \cdot \Mdec \left(\x, \z \right)} + \log \frac{1}{2} (\Menc \left(\x, \z \right) + \Mdec \left(\x, \z \right))} \\
    &= ~ \E{\x,\z \sim \Msamp}{-\log \, \frac{\sqrt{\Menc \left(\x, \z \right) \cdot \Mdec \left(\x, \z \right)}}{\frac{1}{2} (\Menc \left(\x, \z \right) + \Mdec \left(\x, \z \right))} } ~\ge~ 0
\end{align*}
where the inequality follows Jensen's inequality, and equality holds only when $\Menc \left(\x, \z \right) = \Mdec \left(\x, \z \right)$ (\ie,
encoding and decoding distributions are consistent).
In practice we find that encouraging model consistency also helps stabilize learning.

To understand the MIM objective in greater depth, we find it helpful to express $\MIMloss$ 
as a sum of fundamental terms that provide some intuition for its expected behavior.  
In particular, as derived in the supplementary material:
\begin{align}
    \MIMloss(\params)~ &=  ~
    \mathrm{R}_\mathrm{H}(\params) \, +\,
    \frac{1}{4}\big(\, \DKL{\pjoint(\z)}{ \Mdec(\z)}  + \DKL{\pjoint(\x)}{\Menc(\x)} \big)
    \nonumber \\
    & \qquad
    +\, \frac{1}{4}\big(\,\DKL{\Penc }{ \Mdec(\x, \z)} + \DKL{\Pdec}{ \Menc(\z , \x)} \big)
    \label{eq:MIM-parts}
\end{align}
The first term in \eqref{eq:MIM-parts}, as discussed above, encourages high mutual
information between observations and latent states. The second term shows that MIM
directly encourages the model prior distributions to match the anchor distributions.
Indeed, the KL term between the data anchor and the model prior is the maximum
likelihood objective.
The third term encourages consistency between the model distributions and the
anchored distributions, in effect fitting the model decoder to samples drawn from
the anchored encoder (cf.\ VAE), and, via symmetry, fitting the model encoder to
samples drawn from the anchored decoder (both with reparameterization).
In this view, MIM can be seen as simultaneously training and distilling a model
distribution over the data into a latent variable model.
The idea of distilling density models has been used in other domains, e.g.,
for parallelizing auto-regressive models~\citep{oord2017parallel}.

In summary, the MIM loss can be viewed as an upper bound on the entropy
of a particular mixture distribution $\Msamp$:
\begin{eqnarray}
    \MIMloss(\params) &=& 
    ~\frac{1}{2} \big(\, H(\Msamp, \Menc \left(\x, \z \right) ) + 
    H(\Msamp, \Mdec \left(\x, \z \right)) \, \big) \nonumber \\
    &=& H(\Msamp, \Mmodel) + \RMIM(\params) \nonumber  \\
    &\geq& H(\Msamp, \Mmodel) \nonumber \\
    &\geq& H(\Msamp) \nonumber  \\
    &=&  H_{\Msamp} (\x) + H_{\Msamp} (\z) - I_{\Msamp} (\x;\z) 
    \label{eq:MIM-symmetric-bound}
\end{eqnarray}
Through the MIM loss and the introduction of the parameterized model
distribution $\Mmodel$, we are pushing down on the entropy of the anchored
mixture distribution $\Msamp$, which is the sum of marginal entropies minus 
the mutual information.  Minimizing the MIM bound yields consistency of the 
model encoder and decoder, and high mutual information of $\Msamp$ between 
observations and latent states.

\section{MIM in terms of Symmetric KL Divergence}

As given in Equation~\eqref{eq:vae-kl}, the VAE objective can be expressed as minimizing the KL 
divergence between the joint anchored encoding and anchored decoding distributions (\ie, which jointly defines the sample distribution $\Msamp(\x, \z)$). 
Here we refer to $\pjoint(\x)$ and $\pjoint(\z)$ as anchors which are given externally and are not learned.
Below we consider a model formulation using the symmetric KL divergence (SKL),
\begin{align*}
    \mathrm{SKL}(\params) &= 
    \frac{1}{2} \left ( \,\DKL{\Pdec}{\Penc} + \DKL{\Penc}{\Pdec}\, \right )
    ~,
\end{align*}
the second term of which is the VAE objective.

In what follows we explore the relation between SKL, JSD, and MIM.
Recall that the JSD is written as,
\begin{equation*}
    \mathrm{JSD}(\params) ~=~
    \frac{1}{2}\Big( \, \DKL{\Pdec}{\Msamp} + \DKL{\Penc}{\Msamp}\, \Big) ~.
\end{equation*}

Using the identity $\DKL{p}{q} = H(p, q) - H_p(\x, \z)$, we can express the JSD in terms of entropy and cross entropy.
\begin{align*}
    \mathrm{JSD}(\params) &= \frac{1}{2} \bigg ( H(\Pdec, \Msamp) - H_{\Pdec}(\x, \z) \\
    & ~~~~~ + H(\Penc, \Msamp) - H_{\Penc}(\x, \z) \bigg ) \\
    &= \frac{1}{2} \left ( H(\Pdec, \Msamp) + H(\Penc, \Msamp) \right ) - \RH(\params)
\end{align*}
Using Jensen's inequality, we can bound $\mathrm{JSD}(\params)$ from above,
\begin{align}
    \mathrm{JSD}(\params) & \leq \frac{1}{4} \bigg( H_{\Pdec}(\x, \z) + H(\Pdec, \Penc) \nonumber \\
    &~~~~~~ + H_{\Penc}(\x, \z) + H(\Penc, \Pdec)\bigg) - \RH(\params) \label{eq:skl-rh} \\
    &= \frac{1}{4} \bigg( H(\Pdec, \Penc) + H(\Penc, \Pdec) \nonumber \\
    &~~~~~~ + 2\RH(\params) \bigg) - \RH(\params) \nonumber \\
    &= \frac{1}{4} \bigg( \DKL{\Pdec}{\Penc} + \DKL{\Penc}{\Pdec} \nonumber \\
    &~~~~~~ + 4\RH(\params) \bigg) - \RH(\params) \nonumber \\
    &= \frac{1}{4} \left (\DKL{\Pdec}{\Penc} + \DKL{\Penc}{\Pdec} \right ) \nonumber \\
    &= \frac{1}{2}\mathrm{SKL}(\params) \nonumber
\end{align}
If we add the regularizer $\RH(\params)$ and combine terms, we get
\begin{align*}
    \frac{1}{2}\mathrm{SKL}(\params) + \RH(\params) &= \frac{1}{2}\left ( H(\Msamp, \Penc)  + H(\Msamp, \Pdec)\right )
\end{align*}
When the model priors $\Menc(\x)$ and $\Mdec(\z)$ are equal to the fixed priors $\pjoint(\x)$ and $\pjoint(\z)$, this regularized SKL and MIM are equivalent. 
In general, however, the MIM loss is not a bound on the regularized SKL.

In what follows, we derive the exact relationship between JSD and SKL.
\begin{align*}
    \frac{1}{2}\mathrm{SKL}(\params) + \RH(\params) &= \frac{1}{2}\left ( \DKL{\Msamp}{\Penc}  + \DKL{\Msamp}{\Pdec} \right ) + H_{\Msamp}(\x, \z) \\
    &= \frac{1}{2}\left (\DKL{\Msamp}{\Penc} + \DKL{\Msamp}{\Pdec}\right ) \\
    & ~~~~~~ + \mathrm{JSD}(\params) + \RH(\params)
\end{align*}
which gives the exact relation between JSD and SKL.
\begin{align*}
    \frac{1}{2}\mathrm{SKL}(\params) &= \frac{1}{2}\left ( \DKL{\Msamp}{\Penc}  + \DKL{\Msamp}{\Pdec} \right ) + \mathrm{JSD}(\params) \\
    &= \frac{1}{2}\left ( \DKL{\Msamp}{\Penc}  + \DKL{\Msamp}{\Pdec} \right ) \\
    & ~~~~~~+ \frac{1}{2}\left ( \DKL{\Penc}{\Msamp}  + \DKL{\Pdec}{\Msamp} \right )
\end{align*}


\section{Parameterizing \texorpdfstring{$\Menc(\x)$}{q(x)} and \texorpdfstring{$\Mdec(\z)$}{p(z)} for fair comparison with VAEs}
In the MIM framework, there is flexibility in the choice of $\Menc(\x)$ and $\Mdec(\z)$.
To facilitate a direct comparison with VAEs,
we must be careful to keep the architectures consistent and not introduce additional model parameters.
For simplicity and a fair comparison, we set $\Mdec(\z)=\pjoint(\z)$ and leave other considerations for future work.
For $\Menc(\x)$, we consider two different approaches that leverage the decoder distribution $\Mdecoder$. The first is
to consider the marginal decoding distribution,
\begin{align}
    \Menc(\x) &= \mathbb{E}_{\pjoint(\z)}\left [\Mdecoder\right ]. \label{eq:qapprox-mc}
\end{align}
Which we approximate by drawing one sample from $\pjoint(\z)$ when we need to evaluate $\Menc(\x)$. This can suffer from high variance if the prior is far from the true posterior.

The other is to consider an importance sampling estimate, for which we use the encoder distribution $\Mencoder$,
\begin{align}
    \Menc(\x) &= \mathbb{E}_{\Menc(\z|\x)} \left [\frac{\Pdec}{\Menc(\z|\x)} \right ]. \label{eq:qapprox-is}
\end{align}
Where once again, we approximate using one sample, this time from the encoder distribution, and multiply by the importance weight $\frac{\pjoint(\z)}{\Mencoder}$. Samples of $\z$
are drawn using the reparameterization trick~\cite{Kingma2013, Rezende2014} in order to allow for gradient-based training.
We utilize \eqref{eq:qapprox-mc} when sampling from the decoding distribution during the training of a MIM model, and \eqref{eq:qapprox-is} when sampling from the encoding distribution.

\section{Experimentation Details}

Following  \cite{Hjelm2018}, we estimate mutual information 
using the KSG mutual information estimator \cite{PhysRevE.69.066138,DBLP:journals/corr/GaoOV16},  
based on a K-NN neighborhoods with $k=5$, and measure the quality of the representation with classification axuliary task.

The learning algorithm is described in Algorithm\ \ref{algo:vae-as-mim}. In what follows we describe
in details the experimental setup, architecture, and training procedure for the experiments that were presented in the paper.

\begin{figure}[t]
    \centering
    \begin{minipage}[t]{0.95\textwidth}
        \begin{algorithm}[H]
        \caption{\MIM learning with marginal $\Menc(\x)$}
        \label{algo:vae-as-mim}
        \begin{algorithmic}[1]
        \REQUIRE Samples from anchors $\pjoint(\x),\pjoint(\z)$
        \REQUIRE Define $\Menc(\x) = \E{\z \sim \Mdec(\z)}{\Mdec(\x|\z)}$
        \WHILE{not converged}
        \STATE \textcolor{gray}{\textit{\# Sample encoding distribution}}
        \STATE $D_\mathrm{enc} \gets \{ \x_i, \z_i \sim \Menc(\z|\x)\pjoint(\x) \}_{i=1}^{N}$
        \STATE \textcolor{gray}{\textit{\# Compute objective, approximate $\log \Menc(\x)$ with 1 sample}}
        \STATE $\log \Menc(\x_i) \approx \log \Mdec(\x_i|\z_i) + \log \Mdec(\z_i) - \log \Menc(\z_i|\x_i)$
        \STATE $\EMIMloss \left( \params ; D_\mathrm{enc} \right) \gets  -\frac{1}{N}\sum_{i=1}^{N} \left( \log \Mdec(\x_i|\z_i) + \log \Mdec(\z_i) \right)$
        \STATE \textcolor{gray}{\textit{\# Sample decoding distribution}}
        \STATE $D_\mathrm{dec} \gets \{ \x_i, \z_i \sim \Mdec(\x|\z)\pjoint(\z) \}_{i=1}^{N}$
        \STATE \textcolor{gray}{\textit{\# Compute objective, approximate $\log \Menc(\x)$ with 1 sample and importance sampling}}
        \STATE $\log \Menc(\x_i) \approx \log \Mdec(\x_i|\z_i)$
        \STATE $\EMIMloss \left( \params ; D_\mathrm{dec} \right) \gets -\frac{1}{2N} \sum_{i=1}^{N} \left(
                 \log \Menc(\z_i|\x_i) +  2\log \Mdec(\x_i|\z_i) +  \log \Mdec(\z_i) \right)$
        \STATE \textcolor{gray}{\textit{\# Minimize loss}}
        \STATE $\Delta \params \propto -\nabla_{\params} \left( \EMIMloss \left( \params ; D_\mathrm{dec} \right) + \EMIMloss \left( \params ; D_\mathrm{enc} \right) \right)$
        \ENDWHILE
        \end{algorithmic}
        \end{algorithm}
    \end{minipage}
\end{figure}

\subsection{2D Mixture Model Data}

In all experiments we use Adam optimizer \cite{2014arXiv1412.6980K} with $lr = 1e-3$, and
mini-batch of size 128. We stopped training for all experiments when validation loss
has not improved for 10 epochs.

Data are drawn from a Gaussian mixture model with five isotropic components
with standard deviation 0.25, and the latent anchor, $\pjoint(\z)$, is an isotropic standard normal distribution. 
The encoder and decoder are conditional Gaussian distributions, where the means
and variances of which are regressed from the input using two fully 
connected layers and \emph{tanh} activation function.
Following \cite{DBLP:journals/corr/BornscheinSFB15}, the parameterized data prior, $\Menc(\x)$, is defined to be the 
marginal of the decoding distribution, or explicitly $\Menc(\x) ~=~ \E{\z \sim \Mdec(\z)}{\,\Mdec(\x|\z)\,}$, 
where the only model parameters are those of the encoder and decoder, and the encoding distribution $\Menc(\x, \z)$ is defined to be consistent with the decoding distribution $\Mdec(\x, \z)$.
As such we can learn models with MIM and VAE objective that share the same architectures and parameterizations.

\subsection{Representation Learning with MIM in High Dimensional Image Data}

We experiment with convHVAE (L = 2) model from \cite{DBLP:journals/corr/TomczakW17}, with Standard (S) prior which is the usual Normal distributions, and VampPrior (VP) prior which define the prior as a mixture model of the encoder conditioned on learnable pseudo-inputs $\bs{u}_k$, or explicitly $\Mdec(\z) = \frac{1}{K}\sum_{k=1}^{K} \Menc(\z|\bs{u}_k)$. In all the experiments we used the same setup that was used in \cite{DBLP:journals/corr/TomczakW17}, and with the same latent dimensionality $\z \in \mathbb{R}^{80}$. By doing so we aim to highlight the generality of MIM learning as being architecture independent, and to provide examples for the training procedure of existing VAE architectures with MIM learning.

\section{Additional Results}

\subsection{2D Mixture Model Data}

\begin{figure}[ht]
    \centering
    \setlength{\tabcolsep}{0pt}
    \begin{tabular}{*4{>{\centering\arraybackslash}m{0.25\textwidth}}}
      \includegraphics[width=0.24\columnwidth]{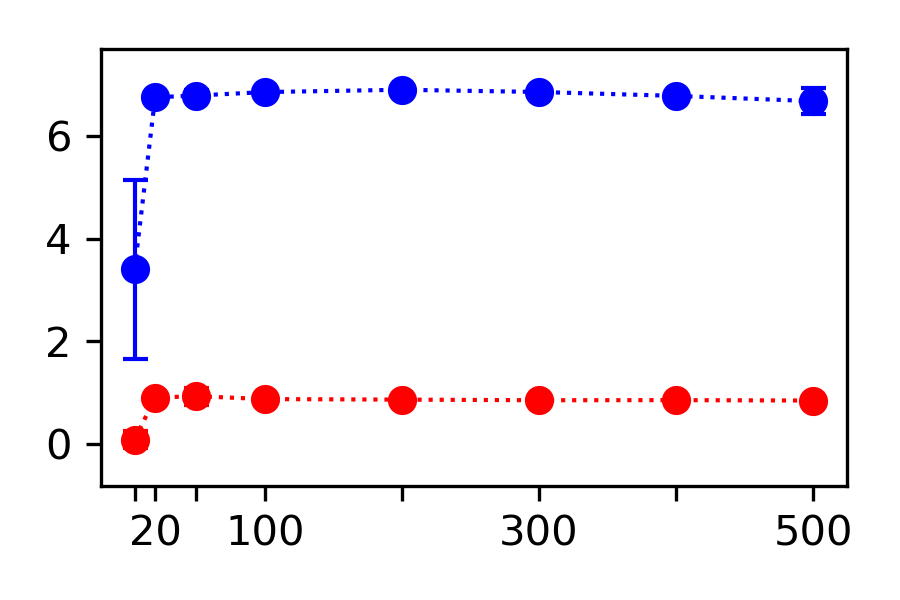}
    & \includegraphics[width=0.24\columnwidth]{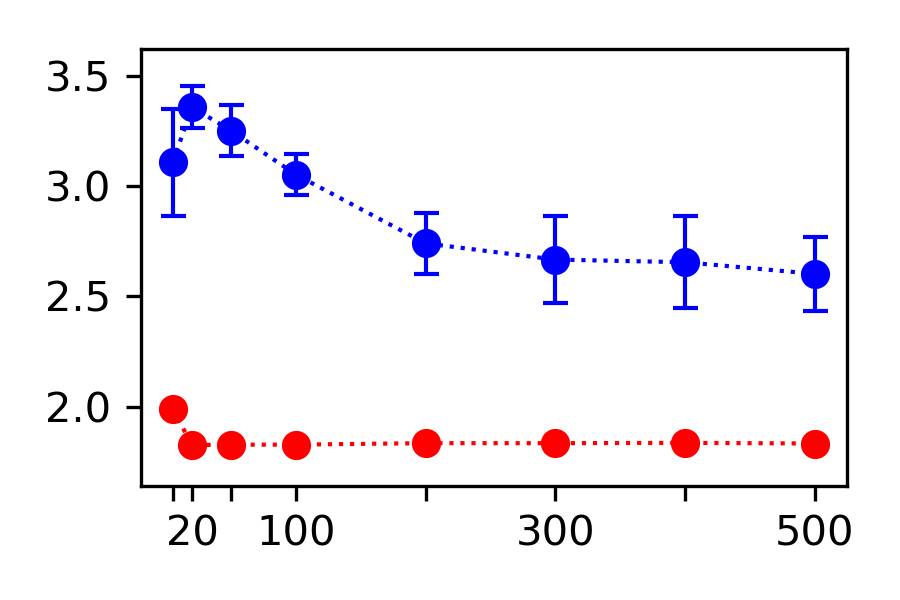}
    & \includegraphics[width=0.24\columnwidth]{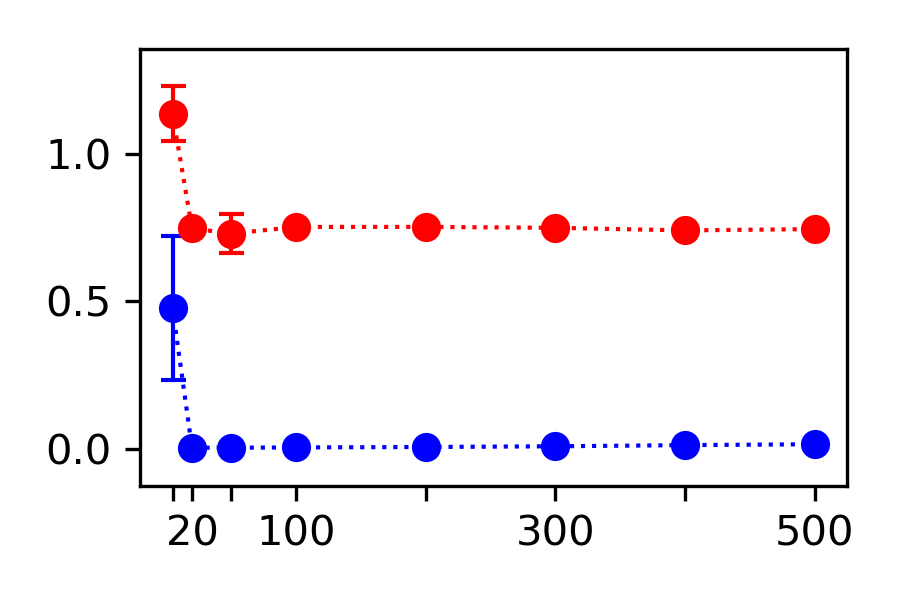}
    & \includegraphics[width=0.24\columnwidth]{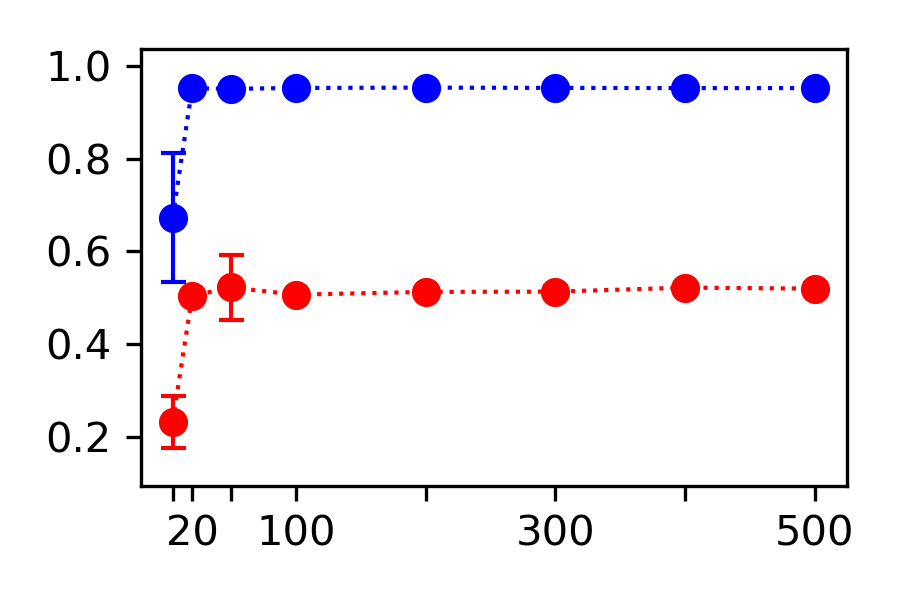}
    \\
    (a) MI & (b)  NLL & (c) Recon.\ Error & (d) Classif.\ (5-NN)
    \end{tabular}
    \caption{Test performance for MIM (blue) and VAE (red) for 2D GMM experiment,
    all as functions of the number of hidden units (on x-axis), based on 10 learned
    models in each case. From left to right, plots show mutual information, log marginal 
    probability of test points, reconstruction error, and k-NN classification performance.
    }\label{fig:posterior-collapse-quantitative}
\end{figure}

Here we quantify the complete experimental results that were presented in
Fig.\ \ref{fig:posterior-collapse-qualitative}.
We plot the mutual information, 
the average log marginal of test points under the model $\Menc$,
the reconstruction error of test points, and 5-NN classification
(predicting which of five GMM components the test points were drawn from).

\subsection{Representation Learning with MIM in High Dimensional Image Data}

\begin{figure}[ht]
    \centering
    \setlength{\tabcolsep}{0pt}
    \begin{tabular}{*2{>{\centering\arraybackslash}m{0.5\textwidth}}}
     \multicolumn{2}{c}{\includegraphics[width=1.0\columnwidth]{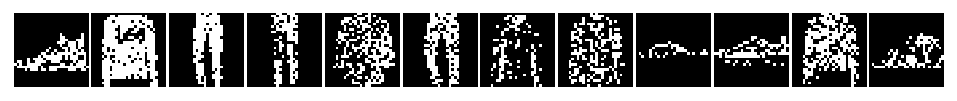}} 
    \\[-0.2cm]
     \multicolumn{2}{c}{\includegraphics[width=1.0\columnwidth]{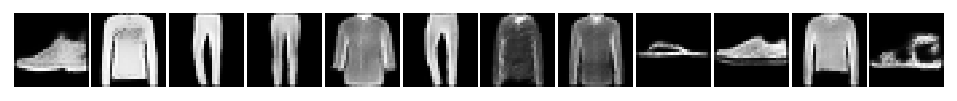}} 
    \\[-0.2cm]
    \multicolumn{2}{c}{\includegraphics[width=1.0\columnwidth]{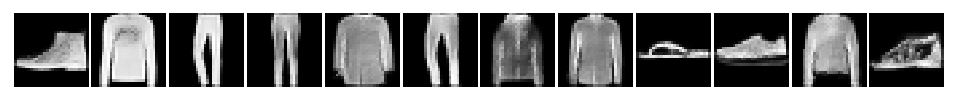}} 
    \\
    \includegraphics[width=0.5\columnwidth]{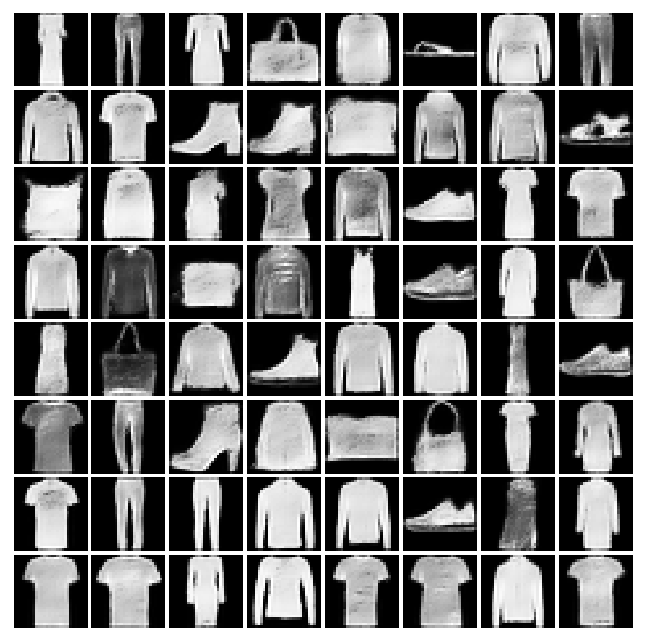}
    & \includegraphics[width=0.5\columnwidth]{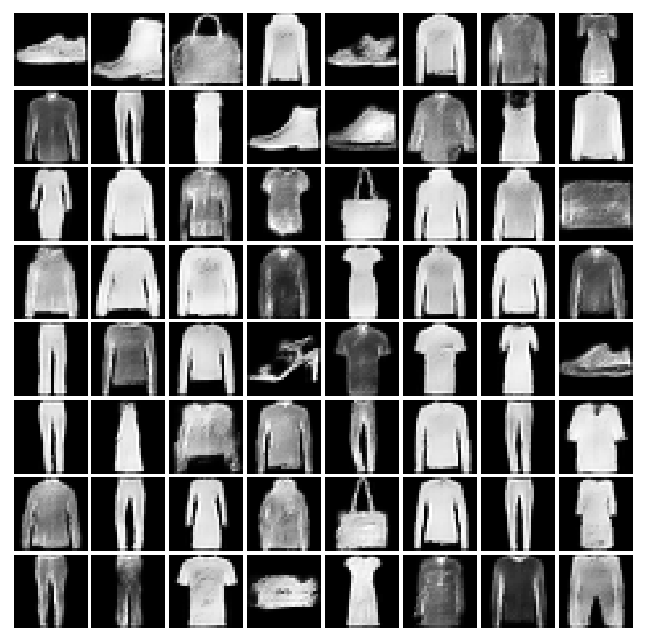}
    \\
    (a) VAE (VP) & (b) A-MIM (VP)
    \end{tabular}
    \caption{VAE and A-MIM learning with PixelHVAE (VP) for Fashion MNIST dataset. Top three rows are data samples, VAE, A-MIM, correspondingly. Bottom row is model samples. }
    \label{fig:mim-vs-vae-image-qualitative-fashion-mnist}
\end{figure}

\begin{figure}[ht]
    \centering
    \setlength{\tabcolsep}{0pt}
    \begin{tabular}{*2{>{\centering\arraybackslash}m{0.5\textwidth}}}
     \multicolumn{2}{c}{\includegraphics[width=1.0\columnwidth]{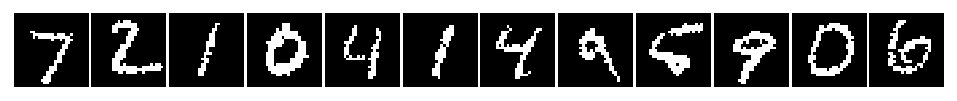}} 
    \\[-0.2cm]
     \multicolumn{2}{c}{\includegraphics[width=1.0\columnwidth]{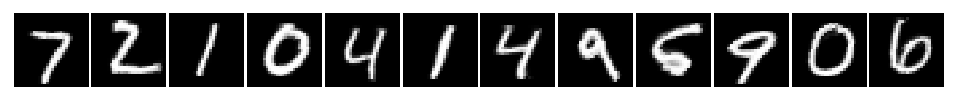}} 
    \\[-0.2cm]
    \multicolumn{2}{c}{\includegraphics[width=1.0\columnwidth]{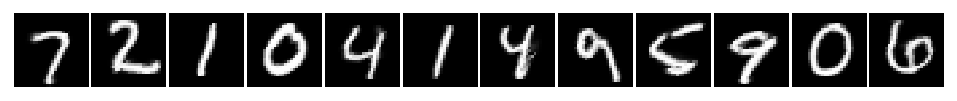}} 
    \\
    \includegraphics[width=0.5\columnwidth]{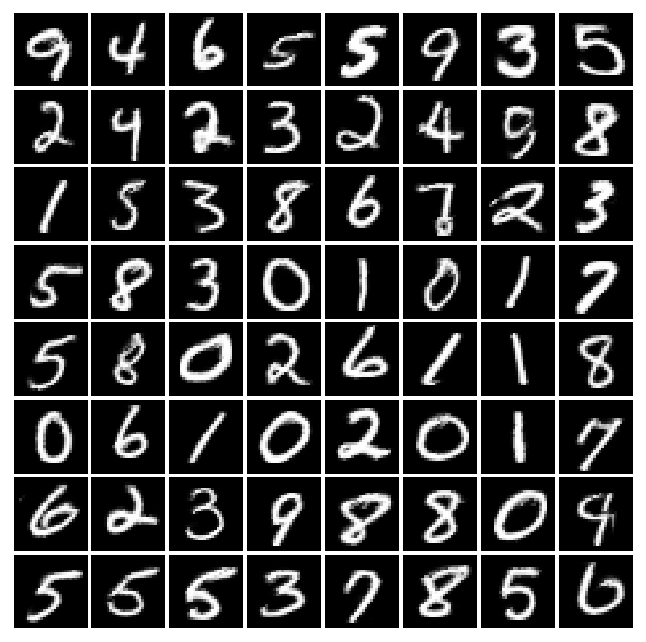}
    & \includegraphics[width=0.5\columnwidth]{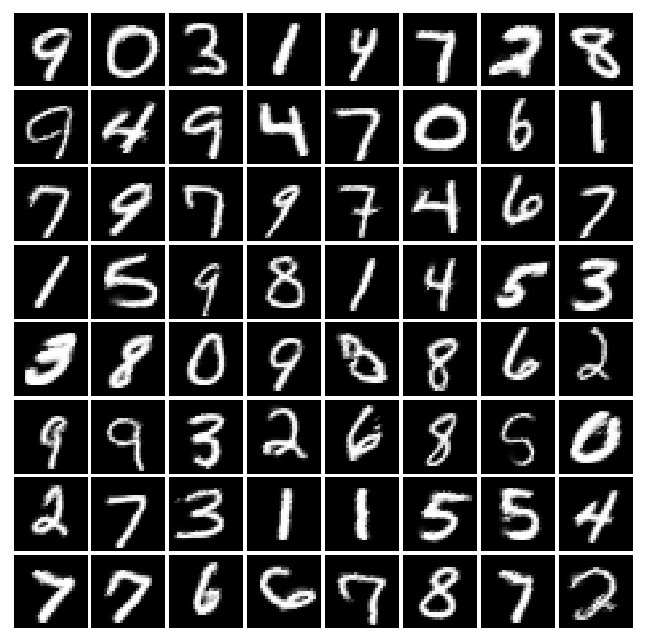}
    \\
    (a) VAE (VP) & (b) A-MIM (VP)
    \end{tabular}
    \caption{VAE and A-MIM learning with PixelHVAE (VP) for MNIST dataset. Top three rows are data samples, VAE, A-MIM, correspondingly. Bottom row is model samples. }
    \label{fig:mim-vs-vae-image-qualitative-mnist}
\end{figure}

Training times of MIM models are comparable to training times for VAEs with 
comparable architectures. The principal difference will be the time required 
for sampling from the decoder during training.  For certain models,
such as auto-regressive decoders \cite{Kingma2016}, this can be significant.  
In such cases (\ie, PixelHVAE here), we  find that we can also learn the model by changing the sampling distribution 
to only include samples from the encoding distribution.
By using asymmetric sampling $\Msamp (\x, \z) = \pjoint(\x) \Menc(\z)$, where we sample from the encoding distribution only (\ie, similar to VAE), training time is comparable to VAE. We name that model A-MIM.

Here we show qualitative results for the most expressive model,  PixelHVAE (VP). Figures\ (\ref{fig:mim-vs-vae-image-qualitative-fashion-mnist}, \ref{fig:mim-vs-vae-image-qualitative-mnist}), depict reconstruction, and sampling for Fashion-MNIST, and MNIST, correspondingly.
The top three rows of each of the plots depicts data samples, VAE reconstruction, and A-MIM reconstruction, respectively. The bottom row depicts samples. 
The results demonstrate comparable samples and reconstruction for MIM and VAE.

\end{document}